\documentclass[11pt]{article}

\usepackage{acl} 

\usepackage{tempora} 
\usepackage{latexsym}
\usepackage[T2A,T1]{fontenc}
\usepackage[utf8]{inputenc}
\usepackage[ukrainian,english]{babel}
\usepackage{microtype}
\usepackage{inconsolata}
\usepackage{graphicx}
\usepackage{booktabs}
\usepackage{multirow}
\usepackage{amsmath}
\usepackage{hyperref}

\title{Balancing Reasoning and Hardware Constraints in RAG Pipelines for Ukrainian Multi-Domain Document Understanding}

\author{Illya Havrylov \\
  National Technical University of Ukraine ``Igor Sikorsky Kyiv Polytechnic Institute'' \\
  Educational and Scientific Institute for Applied System Analysis (IASA) \\
  Department of Artificial Intelligence \\
  \texttt{ilia89279@gmail.com} \\}

\begin{document}
\selectlanguage{english}
\maketitle

\begin{abstract}
This paper describes the system submitted to the UNLP 2026 Shared Task on Multi-Domain Document Understanding. The challenge required extracting precise answers, document IDs, and page numbers from a diverse corpus of Ukrainian PDF documents within a strict 9-hour offline Kaggle execution limit. During evaluation on the hidden private test set, optical character recognition (OCR) of scanned documents emerged as a severe bottleneck, consuming 5--7 hours of the total time budget due to sequential single-threaded execution. This overhead strictly limited the remaining time for Large Language Model (LLM) inference to approximately two hours for 500 questions. To guarantee pipeline completion without timeouts, we developed a resource-efficient Hybrid Retrieval-Augmented Generation (RAG) pipeline utilizing BM25, BGE-M3, and Cross-Encoder reranking. Rather than deploying parameter-heavy reasoning models (e.g., DeepSeek R1) which consistently timed out, we utilized a 4-bit quantized LapaLLM 12B model via \texttt{llama.cpp} on dual NVIDIA T4 GPUs. Prioritizing pipeline stability over multi-step reasoning, our system achieved a Private Score of 0.8095, placing 10th out of 15 active teams.\footnote{Code is available at: \url{https://github.com/catdlia/Notebook_From_UNLP2026}}
\end{abstract}

\section{Introduction}

Multi-domain document understanding presents significant challenges for Ukrainian natural language processing, particularly when parsing digitized legal, medical, and sports documents. The UNLP 2026 Shared Task evaluates AI systems on retrieving information from diverse documents and generalizing across unseen domains.

The competition was hosted on Kaggle in a code-only, offline environment with a strict 9-hour runtime limit. Under these conditions, an unexpected hardware bottleneck emerged: an extensive volume of image-based, scanned PDFs in the hidden private test set required continuous fallback to Tesseract OCR. Because OCR processing was executed sequentially, it consumed 5 to 7 hours of the execution budget, leaving around 2 hours for inference across approximately 500 questions (roughly 14 seconds per query for retrieval and generation combined).

This limitation forced a fundamental engineering trade-off: reasoning depth versus inference latency. While heavy reasoning models theoretically offer higher accuracy, their slow generation speeds and extensive chain-of-thought token footprints caused notebook timeouts.

In this work, we document a pragmatic, domain-agnostic RAG pipeline designed to operate within these constraints. Our main contributions are:
\begin{enumerate}
    \item We deploy a robust dual-stream hybrid retrieval pipeline (sparse BM25 and dense BGE-M3) combined with Reciprocal Rank Fusion (RRF) and Cross-Encoder reranking.
    \item We demonstrate that under extreme runtime constraints, utilizing a 4-bit quantized Ukrainian-adapted model (LapaLLM 12B) on dual NVIDIA T4 GPUs via \texttt{llama.cpp} ensures reliable completion compared to large reasoning architectures.
    \item We show that deterministically decoupling citation extraction (Document ID and Page Number) from LLM generation eliminates attention degradation and ensures stable formatting.
\end{enumerate}

\section{Task and Dataset Description}

\subsection{Dataset Characteristics}
The shared task corpus comprises multi-domain Ukrainian documents across three distinct domains: legal frameworks, medical instructions, and sports regulations. The private test set includes approximately 240 unseen documents formatted as native digital PDFs and scanned image PDFs, introducing substantial noise and layout variations. The test set consists of approximately 500 multiple-choice questions, each providing 6 possible options.

\subsection{Evaluation Metric}
Systems were evaluated on selecting the correct answer option ($a_i$), citing the correct Document ID ($d_i$), and identifying the relevant Page Number ($p_i$). The official evaluation metric is defined as:
\begin{equation}
    \text{Metric} = \frac{0.5}{N} \sum_{i=1}^N a_i + \frac{0.25}{N} \sum_{i=1}^N d_i + \frac{0.25}{N} \sum_{i=1}^N p_i
\end{equation}
where $a_i, d_i \in \{0, 1\}$ are binary indicators of exact matches. The page proximity score $p_i \in$ degrades linearly with distance from the gold page, provided the document ID is correctly identified.

\section{Related Work}

\subsection{Retrieval-Augmented Generation (RAG)}
Standard RAG architectures combine parametric memory from pre-trained language models with non-parametric retrieval from external knowledge bases \citep{rag_lewis}. In complex document understanding, single-retriever systems often struggle with domain-specific terminology. Hybrid retrieval pipelines combining sparse lexical matching (e.g., BM25) and dense neural embeddings via Reciprocal Rank Fusion (RRF) \citep{rrf_cormack} followed by Cross-Encoder reranking have been shown to significantly boost recall and precision across heterogeneous document formats.

\subsection{Ukrainian NLP and Open LLMs}
Natural language processing for Cyrillic and low-resource languages often suffers from poor tokenization efficiency in standard multilingual models. Recently, open Ukrainian language models such as MamayLM \citep{mamaylm} and LapaLLM \citep{lapallm} have introduced expanded vocabularies and targeted pre-training/fine-tuning. This substantially lowers the token-to-word ratio for Ukrainian text, reducing memory footprints and accelerating prefill latency during retrieval augmentation.

\section{System Architecture}

The pipeline comprises three core components: document parsing, hybrid retrieval, and generation.

\subsection{Document Parsing and Chunking}
We utilized \texttt{pymupdf4llm} \citep{pymupdf} to extract per-page text in Markdown format, invoking Tesseract OCR when image scans were encountered. In our submission, OCR was executed sequentially without multiprocessing, which caused the 5--7 hour runtime bottleneck. Standard operating system outputs and C/C++ trace dumps were suppressed to prevent I/O buffer crashes inside the Kaggle environment.

Documents were segmented into fixed lexical windows of 250 words with a 50-word overlap. Empirical checks indicated that semantic chunking frequently fragmented tabular layouts across pages, whereas fixed lexical chunking preserved consistent token density for keyword matching.

\subsection{Hybrid Retrieval Pipeline}
To balance lexical precision and semantic abstraction across diverse domains:
\begin{itemize}
    \item \textbf{Sparse Stream:} \texttt{BM25Okapi} \citep{rank-bm25} retrieved the Top-300 lexical matches from space-tokenized text.
    \item \textbf{Dense Stream:} \texttt{BGE-M3} \citep{bge-m3} generated 1024-dimensional embeddings (mean-pooling, sequence length 512) to retrieve the Top-300 semantic candidates via cosine similarity.
\end{itemize}
The candidate sets were fused using Reciprocal Rank Fusion ($k=60$) and truncated to the Top-100 items. These 100 chunks were rescored using a Cross-Encoder (\texttt{bge-reranker-v2-m3}). The top 7 reranked chunks (capped at 6,500 characters) were formatted into the final LLM context window.

\subsection{Hardware Selection and LLM Inference}
In Kaggle's offline environment, compute options were constrained to either a single NVIDIA Tesla P100 (16GB) or dual NVIDIA T4 GPUs ($2 \times 16\text{GB} = 32\text{GB}$). Although the P100 offers higher FP32 memory bandwidth, its Pascal architecture lacks native support for \texttt{Bfloat16} compute and modern tensor formats. Consequently, we selected the dual NVIDIA T4 environment, which provided 32GB of combined VRAM to comfortably fit larger quantized models and long context windows.

Inference was powered by \texttt{llama.cpp} \citep{llama-cpp} executing a 4-bit quantized (\texttt{Q4\_K\_M}) \texttt{LapaLLM-12B} model \citep{lapallm}. LapaLLM's expanded Cyrillic vocabulary yielded superior token compression for Ukrainian text. Context parameters were configured to \texttt{n\_ctx=6144} and \texttt{n\_batch=1024} with all layers offloaded to GPU (\texttt{n\_gpu\_layers=-1}). 

To maintain reasoning structure, we used a Chain-of-Thought (CoT) prompt instructing the model to analyze context options and output the final answer letter on a designated newline. Metadata citation was completely decoupled: Document ID and Page Number were deterministically assigned from the top-ranked Cross-Encoder chunk, preventing attention shift and hallucinated citations.

\section{Experiments and Results}

\subsection{Leaderboard Results}
As shown in Table~\ref{tab:results}, expanding the context window from 5 to 7 reranked chunks steadily improved retrieval grounding. Our pipeline achieved a Private Score of 0.8095 (10th place out of 15 teams), improving over the Public score (0.7831) and demonstrating strong generalization to unseen domains.

\begin{table}[ht]
\centering
\resizebox{\columnwidth}{!}{%
\begin{tabular}{lccc}
\toprule
\textbf{Configuration} & \textbf{Public} & \textbf{Private} & \textbf{Status} \\
\midrule
MamayLM 7B (Top-5) & 0.6918 & 0.7374 & Completed \\
LapaLLM 12B Base (Top-5) & 0.7741 & 0.7990 & Completed \\
LapaLLM 12B Exp. (Top-7) & \textbf{0.7831} & \textbf{0.8095} & Completed \\
\midrule
LapaLLM 12B Self-Consistency & \textit{Timeout} & \textit{Timeout} & Failed \\
DeepSeek R1 Distill 14B & \textit{Timeout} & \textit{Timeout} & Failed \\
\bottomrule
\end{tabular}%
}
\caption{System performance on the UNLP 2026 leaderboard. Approaches exceeding the 9-hour limit are marked as Timeout.}
\label{tab:results}
\end{table}

\subsection{Ablation and Engineering Bottlenecks}

\paragraph{MamayLM vs. LapaLLM 12B:}
Early iterations with MamayLM exhibited higher hallucination rates and lower tokenization efficiency on dense Ukrainian passages. LapaLLM 12B showed greater extraction fidelity.

\paragraph{Heavy Reasoning Models:}
We evaluated DeepSeek R1-Distill-Qwen-14B \citep{deepseek-r1}. Although it displayed strong deductive ability on training data, its extensive \texttt{<think>} generation tokens, combined with the OCR overhead, consistently triggered execution timeouts.

\paragraph{Self-Consistency \& Multi-Pass Sampling:}
Sampling multiple reasoning paths ($T=0.6, N=10$) for majority voting theoretically enhances robustness but was computationally unfeasible within the remaining 2-hour inference window. Greedy/low-temperature decoding with a single pass was necessary to guarantee completion.

\paragraph{Decoupled Metadata Extraction:}
When prompted to jointly output the text answer and citations (\texttt{Doc\_ID}, \texttt{Page}), the LLM suffered from an ``attention shift'' effect, where formatting constraints degraded question-answering accuracy. Assigning metadata deterministically from the top Cross-Encoder chunk resolved this issue.

\section{Conclusion and Future Work}

This paper presented an efficient RAG pipeline for the UNLP 2026 Shared Task, achieving 10th place under strict 9-hour offline limits. Our results demonstrate that when heavy OCR bottlenecks constrain generation time, combining a dual-stream hybrid retriever with a vocabulary-optimized quantized model (LapaLLM 12B) and decoupled citation assignment provides an effective engineering solution.

In future work, we plan to implement multi-threaded OCR processing to reclaim up to 70\% of the time budget. Furthermore, we aim to investigate custom MLIR compiler architectures and zero-allocation inference runtimes (such as the open-source Tenzo framework \citep{tenzo}) to enable low-latency execution of low-bit quantized reasoning models on edge and compute-constrained hardware.

\section*{Limitations}
The system's design was strictly tailored to Kaggle's 9-hour limit on dual T4 GPUs. The exclusion of heavier reasoning models was driven by runtime constraints rather than fundamental modeling limitations; with parallelized OCR or pre-extracted text, reasoning models would likely achieve higher task accuracy.

\section*{Ethical Considerations}
The dataset contains legal, medical, and sports documents. While the pipeline is optimized for accuracy, automated extraction should not replace human judgment in legal or clinical settings. AI assistants were used for code drafting, debugging, and manuscript proofreading.


\appendix
\section{System Configurations and Hyperparameters}
\label{sec:appendix_params}

Table~\ref{tab:hyperparams} lists the exact hyperparameters used in the final submission pipeline.

\begin{table}[ht]
\centering
\resizebox{\columnwidth}{!}{%
\begin{tabular}{ll}
\toprule
\textbf{Component / Parameter} & \textbf{Value} \\
\midrule
\multicolumn{2}{l}{\textit{Chunking \& Parsing}} \\
Chunk Word Count & 250 words \\
Chunk Overlap & 50 words \\
PDF Parsing Engine & \texttt{pymupdf4llm} (Markdown) \\
OCR Fallback & Tesseract (Single-threaded) \\
\midrule
\multicolumn{2}{l}{\textit{Hybrid Retrieval}} \\
BM25 Candidates & Top-300 \\
Dense Embedding Model & \texttt{BGE-M3} (512 tokens) \\
Dense Candidates & Top-300 \\
RRF Constant ($k$) & 60 \\
RRF Merged Candidates & Top-100 \\
Cross-Encoder Model & \texttt{bge-reranker-v2-m3} \\
LLM Context Chunk Limit & Top-7 (max 6,500 chars) \\
\midrule
\multicolumn{2}{l}{\textit{LLM Inference (\texttt{llama.cpp})}} \\
Model & \texttt{LapaLLM-12B} \\
Quantization & \texttt{Q4\_K\_M} GGUF / \texttt{nf4} \\
Context Size (\texttt{n\_ctx}) & 6144 \\
Batch Size (\texttt{n\_batch}) & 1024 \\
GPU Offload (\texttt{n\_gpu\_layers}) & -1 (All layers offloaded) \\
Sampling Temperature & 0.6 \\
Max Generation Tokens & 1536 \\
Stop Sequences & \texttt{<|im\_end|>}, \texttt{<|eot\_id|>}, \texttt{</s>} \\
Hardware & $2 \times \text{NVIDIA T4 (32GB VRAM)}$ \\
\bottomrule
\end{tabular}%
}
\caption{Hyperparameter configurations extracted from the final submission pipeline.}
\label{tab:hyperparams}
\end{table}

\section{Prompt Templates}
\label{sec:appendix_prompts}

The model was prompted using the ChatML format with structured Chain-of-Thought guidance in Ukrainian.

\paragraph{System Message:}
\begin{center}
\begin{tabular}{|p{0.92\columnwidth}|}
\hline
\small
\textbf{Ukrainian Original:}\\
\foreignlanguage{ukrainian}{Ти елітний ШІ-детектив.}\\[4pt]
\textbf{English Translation:}\\
\textit{You are an elite AI detective.}\\
\hline
\end{tabular}
\end{center}

\paragraph{User Query Template:}
\begin{center}
\begin{tabular}{|p{0.92\columnwidth}|}
\hline
\small
\textbf{Ukrainian Original:}\\
\foreignlanguage{ukrainian}{Контекст:}\newline
\texttt{\{ctx\}}\\[4pt]
\foreignlanguage{ukrainian}{Запитання:} \texttt{\{Question\}}\newline
\foreignlanguage{ukrainian}{Варіанти:}\newline
\texttt{\{options\}}\\[4pt]
\foreignlanguage{ukrainian}{Інструкція: Проаналізуй варіанти на основі контексту. Обов'язково напиши на новому рядку: ``Відповідь: [Літера правильного варіанту]''.}\\[6pt]
\textbf{English Translation:}\\
\textit{Context: \texttt{\{ctx\}}}\newline
\textit{Question: \texttt{\{Question\}}}\newline
\textit{Options: \texttt{\{options\}}}\newline
\textit{Instruction: Analyze the options based on the context. Be sure to write on a new line: ``Answer: [Letter of the correct option]''.}\\
\hline
\end{tabular}
\end{center}

\paragraph{Full ChatML Format:}
\begin{center}
\begin{tabular}{|p{0.92\columnwidth}|}
\hline
\small
\texttt{<|im\_start|>system}\newline
\foreignlanguage{ukrainian}{Ти елітний ШІ-детектив.}\newline
\texttt{<|im\_end|>}\newline
\texttt{<|im\_start|>user}\newline
\foreignlanguage{ukrainian}{Контекст:}\newline
\texttt{\{ctx\}}\\[4pt]
\foreignlanguage{ukrainian}{Запитання:} \texttt{\{Question\}}\newline
\foreignlanguage{ukrainian}{Варіанти:}\newline
\texttt{\{options\}}\\[4pt]
\foreignlanguage{ukrainian}{Інструкція: Проаналізуй варіанти на основі контексту. Обов'язково напиши на новому рядку: ``Відповідь: [Літера правильного варіанту]''.}\newline
\texttt{<|im\_end|>}\newline
\texttt{<|im\_start|>assistant}\\
\hline
\end{tabular}
\end{center}

\end{document}